# IMAGE AND TEXTURE INDEPENDENT DEEP LEARNING NOISE ESTIMATION USING MULTIPLE FRAMES

HIKMET KIRMIZITAŞ [a] AND NURETTIN BEŞLI [a,b]

[a] Harran University, Electrical and Electronics Department Şanlıurfa/Turkey
 *e-mail address*: hkirmizitas@harran.edu.tr

[b] Harran University, Electrical and Electronics Department Şanlıurfa/Turkey
 *e-mail address*: nbesli@harran.edu.tr

Abstract. In this study, a novel multiple-frame based image and texture independent Convolutional Neural Network (CNN) noise estimator is introduced. The estimator works for additive gaussian noise for varying noise levels. The noise level studied in this work has standard deviation equal to 5 to 25 increasing by 5 by 5. Since there isn't any database for noisy multiple images, to train and validate the network user created synthetic noisy images having a variety of noise levels are used by adding additive white gaussian noise (AWGN) to each clean image. For each clean image only 2 frames are created for further calculations. The proposed method is studied on most common gray level images besides the color image databases Kodak, McMaster, BSDS500. Since the proposed method is independent of the noise added as explained later in the paper, further databases were not studied. The image databases constitute of indoor and outdoor scenes having fine details and rich texture. The estimator works with 99 percent accuracy for classification and the scores for the regression are given. The proposed method outperforms the traditional methods. And the regression output can be used with any non-blind denosing method.

## 1. Introduction

Image noise estimation is a critical step for denoising methods. Denoising methods try to reduce the noise in noisy and corrupted images to get a more clear denoisied result. In doing this most of them needs to know the noise level inherent in the noisy image. The success of these algorithms rely on estimating the true noise level added. So research must be done on estimating noise level. Today most of the denoising algorithms are non-blind meaning that they take noise level as a parameter before denoising operation. Among the non-blind methods are Wiener filter[1],Non Local Means[2] and Block-Matching and 3D Filtering(BM3D)[3].CNN is the state-of-the-art image denoising methods[4,5]. There are studies on blind and non-blind versions. Wrong noise level parameter degrades the denoised image quality. So it is a very important step that must be handled carefully. In the literature

∗ OPTIONAL comment concerning the title, *e.g.*, if a variant or an extended abstract of the paper has appeared elsewhere.

thanks, optional.

several noise estimation methods are studied. In the literature among the studies that focus on noise estimation are [6,7,8]. Besides Principal Component Analysis(PCA) based solutions [9] are studied. In [9], smallest eigenvalue of the covariance matrix is chosen. There are mainly 3 types of noise estimation methods. The first is filter based [10], the second is patch based [11-14] and the last is transform-based [15-17] noise estimation method. Filter based methods works for single image and pass the original noisy input image from a high pass filter and the output of the the filter is used to compare with the original noisy image. The difference of two images are taken to decide the noise level. But these approaches suffer from suppressing the image details and smoothing the rich texture and fine details of the clean image. In patch based approaches images are thought to consist of patches which are of size NxN. In patch based methods, the standard deviation of patches are investigated and the one having the least standard deviation among all the patches is taken. The disadvantage of these methods is to overestimate the noise level for images that have small noise levels and underestimate when the image has high noise levels. So in these type of methods the success of estimation depends highly on the inherent noise level. In statistical approaches change in the kurtosis values is affected by the noise corrupted in the image. The estimation when used with denosing methods can increase performance[13]. In recent years, denoising from multiple images and CNNs started to be used. The CNNs are among the state of the art methods for denoising. In CNNs, an architecture of deep learning is trained with noisy images and corresponding noise-free clean images as inputs and outputs to the deep network respectively. In [18], the noise is detected in the image and noise level is classified in 10 bins. In this study if the image does not have noise it is classified as noise free. Apart from noise-free class there are 9 more classes having standard deviation equal to 10 to 90 increasing 10 by 10. The CNN architecture is MatConvNet having 4 convolutional layers,2 max-pooling layers,1 Relu layer and finally a softmax layer for classification purposes. In this study noisy images are classified implicitly putting a noisy image to a class. Noise level is not calculated explicitly so it cannot be used from other non-blind denoising methods. In [19], noise is estimated pixelwise. Since real world noise is different from synthetically corrupted noisy images, instead of a global scalar noise level, every pixels' noise level is estimated using deep learning. It is a successful work surpassing mostly the state-of-the-art methods Liu[20], Pyatykh[9], and Chen[21]. They use a stack of residual patches. In this residuals patches there are no pooling or interpolation operation. But still the noise estimation results can be improved. In [22], noise level is proposed to be estimated using Singular Value Decomposition(SVD) and neural network. The tail parts of the singular values of an image increases with increasing noise level constituting a measure of noise level. So they use this singular values as inputs to the neural network. They train the network with singular values as the inputs to the network and noise's standard deviation as the output of the network. In this study they estimate different noise levels with higher speed and more precisely for both gaussian and hybrid noise.

## 2. SQUEEZENET

Squeezenet is a small version of Alexnet providing three advantages. First of all it requires less communication between servers while training. Secondly it requires less bandwith while transferring network data. For example on Tesla's autopilot semi-automatic cars transferring network data diminishes from 250 MB to 0.5 MB. Thirdly, they are more suitable for hardware implementation. Squeezenet has one fifth less parameters as compared to Alexnet with sufficient accuracy. Further,it requires one hundredth less storage space when compressed. In Squeezenet different from other deep learning architecures a fire module is defined. And they construct the whole architecture with these fire modules. A typical fire module consists of following layers: A squeeze convolution layer having only 1x1 layers before an expand layer which has a combination of 1x1 and 3x3 filters as shown in Fig 1.In this step 1 ninth of less storage occupation is achieved because 1x1 filters have 1 ninth space compared to 3x3 filters. After that they describe three tunable dimensions: s1x1, e1x1 and e3x3. s1x1 is the count of filters in the squeeze layer all of which are 1x1, e1x1 is the count of 1x1 filters in the expand layer as described in Fig 1. And e3x3 is the count of layers having size 3x3 in the expand layer as shown in Fig 1. Second decreasing of dimension occurs by setting s1x1 to be less than (e1x1+e3x3).Thus they decrease the count of input channels to 3x3 filters.The layers of the network of Squeezenet is shown in Figure 2.

## 3. PROPOSED METHOD

In the proposed method, squeezenet deep learning architecture is used for regression. The last two layers of the SquuezeNet architecture is replaced with a fullyconnected layer and a regression layer. The proposed method consists of CNN training and testing steps. In training phase, taken an image, 2 noisy images of standard deviation are created. Each image has 2 noisy images. These two noisy images are created with the assumption that the original image is noise free and secondly the noise added is purely additive white gaussian noise. As it will be explained later in the paper, the noise added must be additive rather than multiplicative or any other style of noise. If the noise added to both frames is purely additive , this proposed method gives image and texture independent noise estimation results. In the input layer of the CNN network, the difference image by subtracting these 2 noisy images is fed to the neural network. As can be seen from Figure 3, the input of the network is just noise.The output of the network is the standard deviation estimate ranging from 5 to 25.After the network is trained with 2000 samples for each standard deviation i.e ranging from 5 to 25,we go to the testing step. In the testing step the difference of the 2 noisy images are calculated and fed to the network. In gray level images the input layer has one channel and in the colour images the network has 3-channel (RGB) image residuals calculated from multiple frames(here taken as 2).

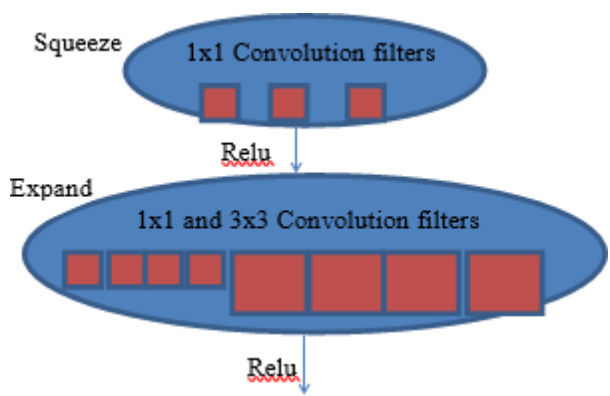

Fig 1.Detailed description of fire modules of Squeezenet

## 4. Results

As a result of the algorithm, the below table is obtained. In the Table 1. the outputs of 4 existing methods which are among the state-of-the-art noise estimation methods till 2022 and lastly the results of the proposed method is given. In this study, we took 2000 images for training and 500 images for validation for the CNN. In classification of noise with 5 bins (noise level 5-10-15-20-25 ), the proposed method has an accuracy rate of 100 percent. For Kodak database there are only 24 images in the database. For McMaster database there 18 images in the database. For BSDS500 database there are 500 images in the database. Some of the images in BSDS500 database has details and fine texture. For each image in the database we constructed 2000 noisy versions of a single clean image and used these images while training.The regression training results are shown in Figure 4. Since the proposed method is independent of the image and texture, we didn't try it on different databases as explained below. As can be seen Figure 5,no matter the input image,the network input is just the noise. After taking the difference of two noisy images the input becomes purely the difference of two randomly distributed noise signals. As a result the input image becomes unimportant for this method. So any other images from other datasets was not tried. For the CNN architecture as explained above, CNN consists of fire modules. For the normal squeeze net architecture first there is a conv1 layer after that comes a maxpool operation and fire module. After 3 fire modules at this step, there is maxpool operation and there are 4 fire modules respectively. After this 4 fire modules, there is a maxpool operation and 1 fire module. After that there is a convolution layer and a regression layer. The noisy images are created synthetically by adding AWGN with Matlab's imnoise function. In application creating images synthetically by adding AWGN to noise free images twice means that two noisy pictures of a reference scene is taken consecutively having zero delay in time. In practice we took the frames from Lenovo K5 cellular phone consecutively having 0.3 ms delay in time. The camera was fixed so any motion blur due to camera's position did not occur but in time if there were motion in the scene it would create blur artifact in the picture. To diminish this possibility of creating motion artifact the time gap between two consecutive frames must be held low as low as possible. And we could not keep this gap below 0.3 seconds in our images due to the software of the cellular phone. The images were similar in pixel so we again added AWGN to the real image taken from cellular phone and obtained the same results as in Table 1. The results of the proposed method are compared with the state-of-the-art methods Tan[2], Chen [21], Liu[20], Pyatykh's[9] in noise estimation. The source code of these implementations can be downloaded from the internet. For colour images the three channels are treated independently and as the final result the average of three channels are calculated. The results of all the methods can be found Table 1. As seen in Table 1, the regression network is trained with 5 levels of AWGN with sigma = 5,10,15,20,25. The proposed method wins 11 first places out of 15 and one second place out of 15.

## 5. EXPERIMENTS

Fully Convolutional networks (FCNs) work good when correspondence of images with image and/or noise is necessary. But while using FCNs there is a point that must be taken care of. It is that the number of convolutional layers must be chosen according to its generalization capability. If the number of layers is chosen too small or too large the network can converge to an undesirable point. Here we used 65 layers. But we limit the number of epochs to 70 so as not to memorize the noise and corresponding output. Too few layers as 10 couldn't be successful. Most of the denoising CNN's consist of many convolutional layers but this cause the result to lose fine details so we used pooling and relu layer as in sequeezenet.

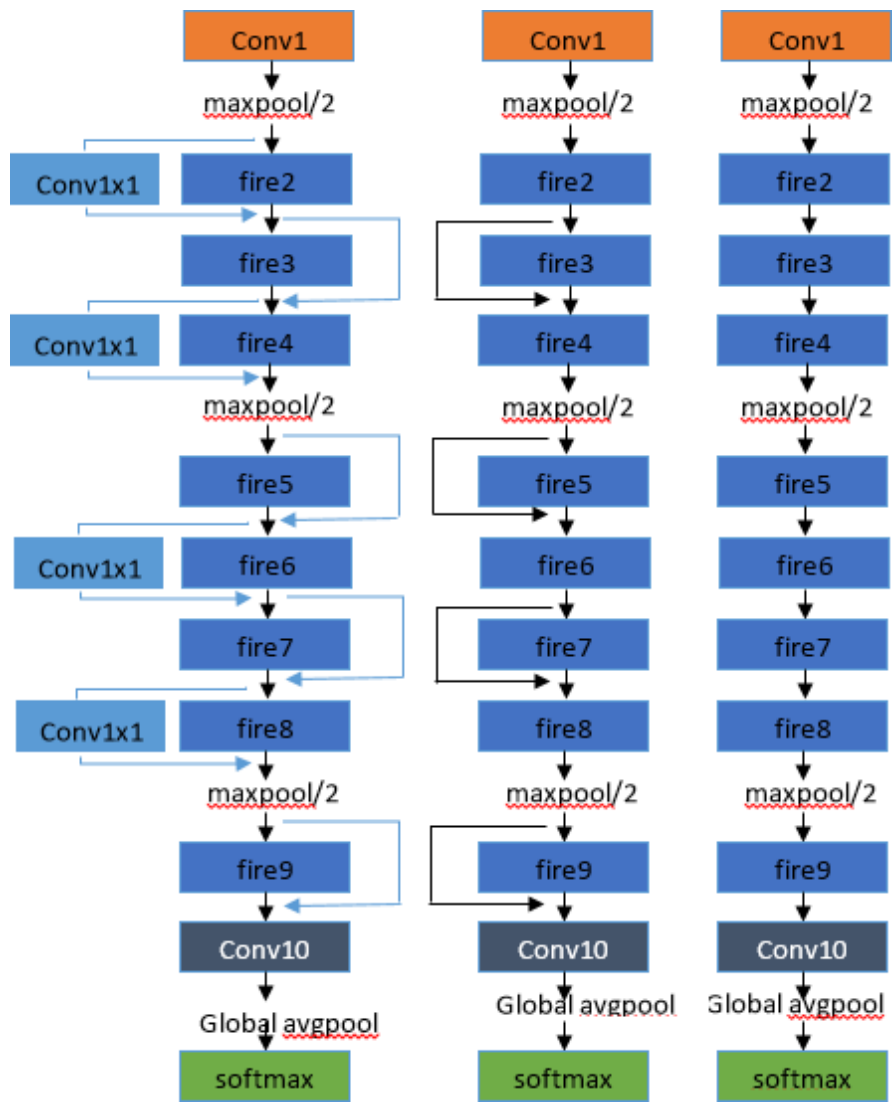

Fig. 2. Architecture of SqueezeNet Network. From left to rigt, SqueezeNet , Network with simple bypass ;Network with complex bypass.

Our results mostly rely on the input we give to the CNN. The difference of two noisy images is good to characterize the noise inherent in the image. In Fig. 4 the prediction of the trained network can be seen. For small values of the regression value the error is too less both in value and percentage. When the regression value increases both the regression value and percentage of the error increases. The classification system's solver was 'adam' and the initial learning rate was chosen to be 0.01. The validation frequency of the training was 50.The number of epochs was chosen to be 30 but in 3 epochs the network was successful to get accuracy of 100 percent.Minimum batch size was 128.Execution environment was gpu. Since there was 4500 samples per class totally 112500 samples, s too much for ordinary cpu, gpu was selected as training environment. L2Regularizaiton was 0.0001 and the gradient threshold method was L2Norm.  Gradient threshold and validation patience was chosen to be infinity. The data was shuffled at every epoch yielding a more robust learning. Learn rate schedule was none and learn rate drop factor was 0.1. Learn rate drop period was 10 and the momentum was chosen to be 10. The results can be seen on Fig.5.As the epoch number increases the accuracy of the trained network increases. The system in Fig. 5 tries to classify 25 noise levels i.e it has 25 different levels starting from 1 to 25 increasing 1 by 1.When the number of classes was 5 with noise level from 5 to 25 increasing 5 by 5,the classification accuracy is 100 percent as in the system classes increasing one by one till 25.The classification accuracy is too high because the system could learn the difference image well. And in classification the samples are forced to be one of the output classes. In regression there is also a regression force which forces the output to be one of the regression levels but this is not as much as in the classification process.

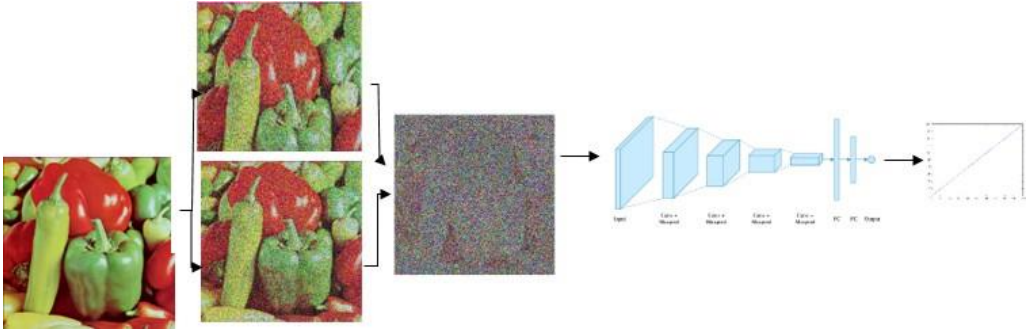

Fig. 3. The proposed method schematic.

Table 1: Noise estimation results on different datasets

| Dataset | Noise Level | Pyatkh[9] | Liu[20] | Chen[21] | Tan[19] | Proposed Method |
|---|---|---|---|---|---|---|
| Kodak(24 Images) | $\sigma = 5$ | 0.51 | 0.26 | **0.15** | **0.15** | 0.23 |
| | $\sigma = 10$ | 1.00 | 0.49 | 0.36 | 0.30 | **0.14** |
| | $\sigma = 15$ | 1.46 | 0.67 | 0.55 | 0.50 | **0.28** |
| | $\sigma = 20$ | 1.91 | 0.86 | 0.75 | 0.71 | **0.28** |
| | $\sigma = 25$ | 2.31 | 1.07 | 0.95 | 0.96 | **0.01** |
| McMaster(18 Images) | $\sigma = 5$ | 0.17 | 0.18 | 0.28 | **0.11** | 0.23 |
| | $\sigma = 10$ | 0.52 | 0.46 | 0.34 | 0.32 | **0.14** |
| | $\sigma = 15$ | 0.87 | 0.76 | 0.67 | 0.62 | **0.28** |
| | $\sigma = 20$ | 1.11 | 1.09 | 1.03 | 0.94 | **0.28** |
| | $\sigma = 25$ | 1.35 | 1.42 | 1.37 | 1.28 | **0.01** |
| BSD500(500 Images) | $\sigma = 5$ | 1.12 | 1.31 | **0.08** | 0.11 | 0.23 |
| | $\sigma = 10$ | 0.80 | 0.42 | 0.22 | 0.26 | **0.14** |
| | $\sigma = 15$ | 1.25 | 0.61 | 0.41 | 0.47 | **0.28** |
| | $\sigma = 20$ | 1.71 | 0.85 | 0.64 | 0.72 | **0.28** |
| | $\sigma = 25$ | 2.17 | 1.11 | 0.91 | 1.00 | **0.01** |

Table 2: Noise estimation execution times in seconds

| Database | Pyatykh[9] (CPU) | Liu[20] (CPU) | Chen[21] (CPU) | Tan[19](CPU/GPU) | Proposed Method |
|---|---|---|---|---|---|
| McMaster | 1.75 | 2.20 | 0.27 | 5.31/1.16 | 0.01 |

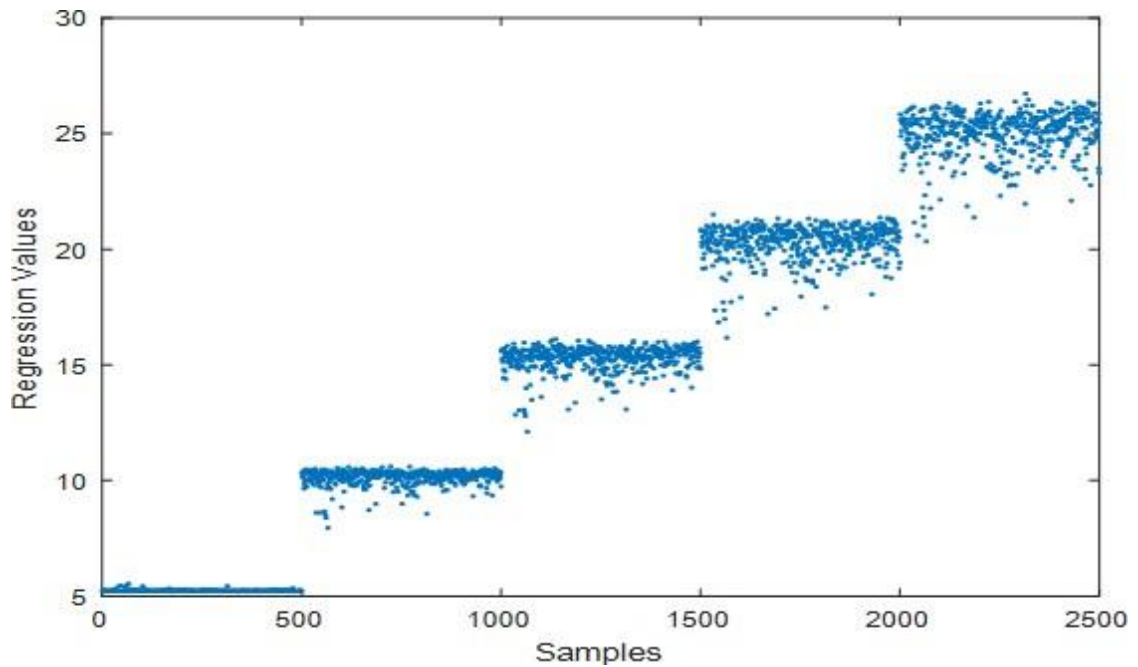

Fig. 4. The Regression plot validation data

Our input works independent of the original picture. Most of the methods studied show that if the image has fine details the success of the noise estimation decreases. In their simulations and results they mention this. But in our case no matter how much the original image has fine details, the proposed noise estimation algorithm works with same accuracy. If we denote it mathematically we can say that if the first noisy image is I+1 and the second image as I + 2, then the difference is I+1 - I+2 yielding 1 – 2. Here the original pixel value disappears since I's cancel each other. So the training input we give to the network becomes independent of the image and texture. For the realization of training for classification and regression we used sequuezenet as it is. The only difference was on the last two layers for regression. For the last two layers we took the original softmax and classification layer, instead we put fullyconnected and regression layers. 'Adam' was used for training and the training was regression instead of classification. So our method can be used by other non-blind denoising methods. We took 2000 samples for each regression value (i.e = 10, 20, 30, 40, 50) and totally we get 10000 training samples. The result metric was average error. The test datasets were Kodak, McMaster and BSDS500.When comparing the results other parameters were kept same. The aim of this paper is to serve noise level truly for denoising applications. Especially non-blind applications take ground-truth noise level as an input parameter to their solutions.

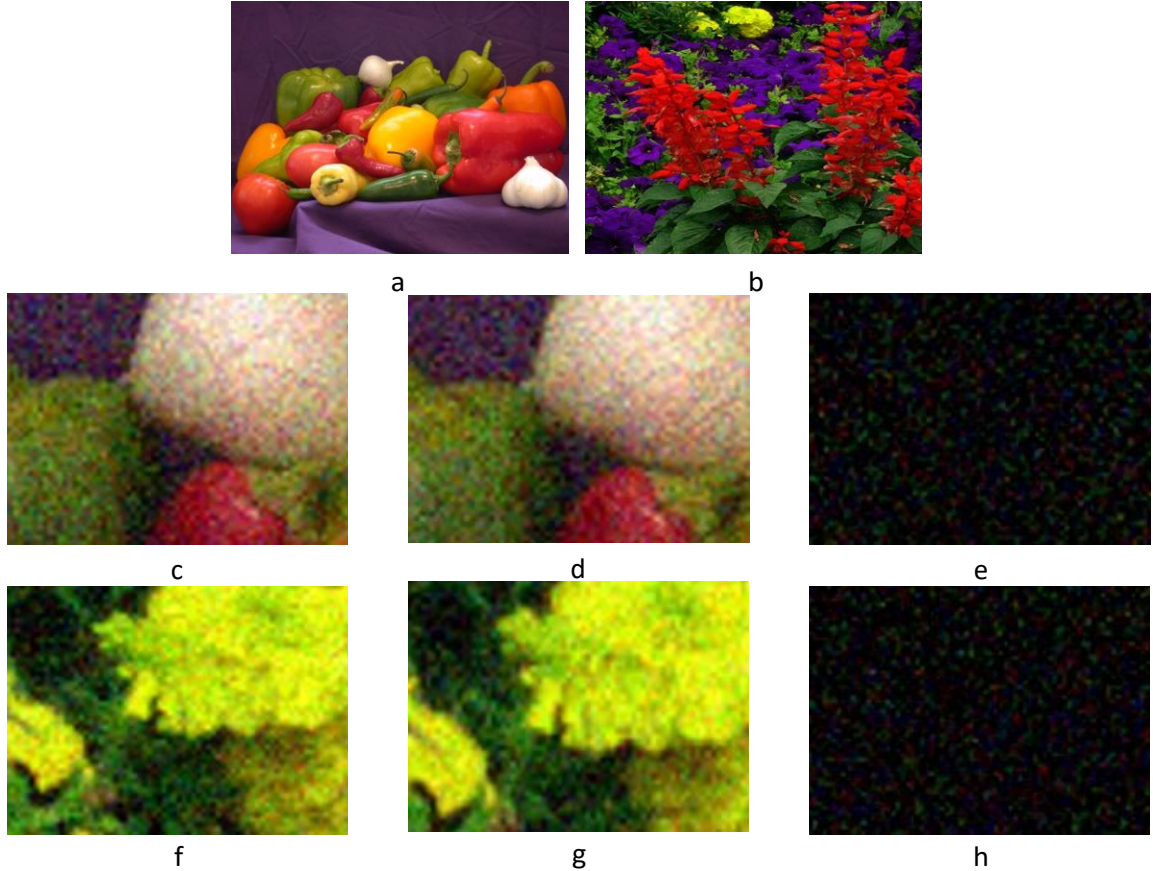

Figure 5: Original peppers image(a) and an image taken from Kodak Dataset(b), c,d,f and g are noisy and enlarged images taken from a and b ,e and h are the difference network inputs.

## 6. CONCLUSION

In this paper, we propose a gaussian noise estimation method which outperforms the state of the art noise estimation methods. By its architecture it can be used by several non-blind denoising methods.It means that Squeezenet is successful to learn the difference of 2 noisy images as explained in the study. The input of the network is purely the noise added to the original noise free images. No matter how the original image is the system can learn it and do classification and regression.Both for classification and regression,the results in Table 1 and regression plot of the network in Figure 4 supports this behavior. This is a big advantage because all of the noise estimation methods behave bad when the original image has details and fine texture. The fine texture hardens the method to estimate it. As can be seen from the results the proposed method is independent of image and texture, i.e. it works with the same accuracy for all of the images as long as the noisy images are created by the method explained in this paper. The results show that the noise level of the difference of 2 frames is liable to be predicted by a CNN. As compared to other state-of-the-art noise estimation methods our proposed method works better.

## REFERENCES

[1] N. Wiener. Extrapolation, Interpolation, and Smoothing of Stationary Time Series, *Cambridge,Technology Press of Massachusetts Institute of Technology*, 1st ed., vol. 7. New York,Wiley, USA: MIT press Cambridge,pp. 4 , 1949. DOI: https://doi.org/10.7551/mitpress/2946.001.0001

[2] A. Buades, B. Coll and J. M. Morel. A non-local algorithm for image denoising, *In Proceedings of the IEEE Computer Society Conference on Computer Vision and Pattern Recognition (CVPR 2005)* San Diego, CA, USA, vol. 2, pp. 60–65, 2005, DOI: 10.1109/CVPR.2005.38

[3] K. Dabov,A. Foi, V. Katkovnik and K. Egiazarian, Image denoising with block-matching and 3D filtering, *in Proceedings Volume 6064, Image Processing: Algorithms and Systems, Neural Networks, and Machine Learning* San Jose, CA, USA, vol. 6064, p. 606414, 2006,DOI: https://doi.org/10.1117/12.643267

[4] H. C. Burger, C. J. Schuler and S. Harmeling, Image denoising: Can plain neural networks compete with BM3D?, *in 2012 IEEE Conference on Computer Vision and Pattern Recognition*, pp. 2392- 2399, 2012,DOI: 10.1109/CVPR.2012.6247952

[5] K. Zhang, W. Zuo, Y. Chen, D. Meng and L. Zhang, Beyond a gaussian denoiser: Residual learning of deep CNN for image denoising, *IEEE Transactions on Image Processing : a Publication of the IEEE Signal Processing Society,*vol. 26,no. 7,pp. 3142-3155. ,2017,DOI: https://doi.org/10.48550/arXiv.1608.03981

[6] P. Jiang and J. Z. Zhang, Fast and reliable noise level estimation based on local statistic, *Pattern Recognition Letters*, vol 78,pp.8 – 13, 2016, https://doi.org/10.1016/j.patrec.2016.03.026

[7] C. Liu, W. T. Freeman, R. Szeliski and S. B. Kang, Noise estimation from a single image, *2006 IEEE Computer Society Conference on Computer Vision and Pattern Recognition (CVPR'06)*, vol. 1, pp. 901–908, 2006, doi: 10.1109/CVPR.2006.207.,DOI: 10.1109/CVPR.2006.207

[8] H. Yue, J. Liu, J. Yang, T. Nguyen and C. Hou, Image noise estimation and removal considering the bayer pattern of noise variance, *In 2017 IEEE International Conference on Image Processing* (ICIP, pp. 2976-2980 ), 2017,DOI: 10.1109/ICIP.2017.8296828

[9] S. Pyatykh, J. Hesser and L. Zheng, Image noise level estimation by principal component analysis, *in IEEE Transactions on Image Processing*, vol. 22, no. 2, pp. 687-699, Feb. 2013,DOI: 10.1109/TIP.2012.2221728

[10] S. K. Abramov, V. V. Lukin, B. Vozel, K. Chehdi and J. T. Astola, Segmentation-based method for blind evaluation of noise variance in images, *Journal of Applied Remote Sensing*, vol. 2,no. 1,pp. 1-16,1 August 2008,DOI: https://doi.org/10.1117/1.2977788

[11] D. H. Shin, R. H. Park, S. Yang and J.H. Jung, Block-based noise estimation using adaptive Gaussian filtering, *IEEE Transactions on Consumer Electronics*, vol. 51, no. 1, pp. 218-226, Feb. 2005,DOI: 10.1109/TCE.2005.1405723

[12] A. Amer and E. Dubois, Reliable and fast structure oriented video noise estimation, *in IEEE Transactions on Circuits and Systems for Video Technology*, vol. 15, no. 1, pp. 113-118, Jan. 2005,DOI: 10.1109/ICIP.2002.1038156

[13] D. H. Shin, R.H. Park, S. Yang and J.H. Jung, Block-based noise estimation using adaptive Gaussian filtering, *in IEEE Transactions on Consumer Electronics*, vol. 51, no. 1, pp. 218-226, Feb. 2005,DOI: 10.1109/TCE.2005.1405723

[14] C. H. Wu and H. H. Chang, Superpixel-based image noise variance estimation with local statistical assessment, *EURASIP Journal on Image and Video Processing*,vol. 2015,no. 1,pp. 38, 2015,DOI: 10.1186/s13640-015-0093-2

[15] S. M. Yang and S. C. Tai, Fast and reliable image-noise estimation using a hybrid approach, *Journal of Electronic Imaging*, vol. 19,no 3,pp. 033007-033007, July 2010,DOI: https://doi.org/10.1117/1.3476329

[16] S. Gai, G. Yang, M. Wan and L. Wang, Hidden markov tree model of images using quaternion wavelet transform, *Computers Electrical Engineering*, vol. 40, no. 3, pp. 819–832, 2014,DOI: 10.1016/j.compeleceng.2014.02.009

[17] M. Hashemi and S. Beheshti, Adaptive noise variance estimation in bayesshrink, *in IEEE Signal Processing Letters*, vol. 17, no. 1, pp. 12–15, Jan. 2010,DOI: 10.1109/LSP.2009.2030856

[18]J. H. Chuah, H. Y. Khaw, F. C. Soon and C. Chow, Detection of Gaussian noise and its level using deep convolutional neural network, *in TENCON 2017 - 2017 IEEE Region 10 Conference*, pp. 2447-2450, 2017,DOI: 10.1109/TENCON.2017.8228272

[19]H. Tan, H. Xiao, S. Lai, Y. Liu and M. Zhang, Pixelwise estimation of signal-dependent image noise using deep residual learning, *Computational Intelligence and Neuroscience*, vol. 2019, pp. 4970508:4970520, 2019.,DOI: https://doi.org/10.1155/2019/4970508

[20] X. Liu, M. Tanaka and M. Okutomi, Single-image noise level estimation for blind denoising, *in IEEE Transactions on Image Processing*, vol. 22, no. 12, pp. 5226-5237, Dec. 2013,DOI: 10.1109/TIP.2013.2283400

[21] G. Chen, F. Zhu and P. A. Heng, An efficient statistical method for image noise level estimation, *in 2015 IEEE International Conference on Computer Vision (ICCV)*, Santiago, Chile, pp. 477–485, December 2015,DOI: 10.1109/ICCV.2015.62

[22] Z. Wang and G. Yuan, Image noise level estimation by neural networks, *In Proceedings of the 2015 International Conference on Materials Engineering and Information Technology Applications*, pp. 692-697, 2015/08,DOI: 10.2991/meita-15.2015.126

[23] F. N. Iandola, M. W. Moskewicz, K. Ashraf, S. Han, W. Dally et al. SqueezeNet: AlexNet-level accuracy with 50x fewer parameters and¡ 0.5 MB model size *arXiv*,vol. abs/1602.07360,DOI:https://doi.org/10.48550/arXiv.1602.07360